# Fully-Convolutional Intensive Feature Flow Neural Network for Text Recognition

**Zhao Zhang**[1,2] and **Zemin Tang**[1] and **Zheng Zhang**[3] and **Yang Wang**[2] and **Jie Qin**[4] and **Meng Wang**[2]

**Abstract.** The Deep Convolutional Neural Networks (CNNs) have obtained a great success for pattern recognition, such as recognizing the texts in images. But existing CNNs based frameworks still have several drawbacks: 1) the traditaional pooling operation may lose important feature information and is unlearnable; 2) the traditional convolution operation optimizes slowly and the hierarchical features from different layers are not fully utilized. In this work, we address these problems by developing a novel deep network model called Fully-Convolutional Intensive Feature Flow Neural Network (IntensiveNet). Specifically, we design a further dense block called intensive block to extract the feature information, where the original inputs and two dense blocks are connected tightly. To encode data appropriately, we present the concepts of dense fusion block and further dense fusion operations for our new intensive block. By adding short connections to different layers, the feature flow and coupling between layers are enhanced. We also replace the traditional convolution by depthwise separable convolution to make the operation efficient. To prevent important feature information being lost to a certain extent, we use a convolution operation with stride 2 to replace the original pooling operation in the customary transition layers. The recognition results on large-scale Chinese string and MNIST datasets show that our IntensiveNet can deliver enhanced recognition results, compared with other related deep models.

## 1  INTRODUCTION

Optical character recognition (OCR) [6] is an important topic in the areas of pattern recognition, which aims at recognizing the texts, including characters and numbers, in images [36]. Although OCR has been extensively studied for last decades, recognizing the texts from natural images accurately is still a challenging task due to the complicated background and contents in the images [2][49][50]. In fact, a character/number may have different appearances in different images because of style, font, resolution or illumination changes. In recent years, with the increasing breakthroughs in areas of computer vision [12-14] and deep learning [7-11][46][51], certain advanced end-to-end text recognition frameworks have been developed [24], which includes the complex two-step pipelines. The first step is to detect the regions of texts in images and the second one is to recognize the textual contents of identified regions [24]. Following this pipeline, some deep network models, such as STN-OCR, are proposed for the text recognition [16]. STN-OCR jointly learns a spatial transformer network [16] to detect text regions in an image, and one text recognition network to recognize the textual contents within identified regions. However, it is not easy to train STN-OCR in practice, and this model cannot fully detect the texts in arbitrary locations in images [24]. Arbitrary orientation network (AON) is recently proposed to recognize the oriented texts arbitrarily and obtain impressing results on irregular and regular texts from images [2]. To be specific, AON presented an arbitrary orientation network to extract the visual features of characters from four directions, employs a filter gate mechanism for feature fusion, and uses an attention-based decoder to generate the character sequences [2]. Recently, a popular framework referred to as Convolutional Recurrent Neural Network (CRNN) [29] has been proposed, which integrates the advantages of Convolutional Neural Networks (CNNs) and Recurrent Neural Networks (RNNs) [18-20]. Although lots of CRNN based deep networks have been derived for text recognition, recent works also reveal that even without the recurrent layers, the reduced models can still obtain promising recognition results with higher efficiency [47]. As such, the framework of CNN + Connectionist Temporal Classification (CTC) [1] is a promising efficient solution. Note that for the original images, CRNN needs to utilize the Connectionist Text Proposal Network (CTPN) [4] to detect the text lines, the outputs of which are the inputs of CRNN.

It is noteworthy to point out that CRNN based frameworks can use the Dense Convolutional Network (DenseNet) [5] as a feature extractor and drop the recurrent layers for efficiency, which forms a new framework termed DenseNet + CTC [48]. But this series of methods still have several disadvantages. First, mainstream CNNs based models usually use pooling as the down-sampling operation to reduce the size of features, but the pooling operation is unlearnable and may be fragile to lose important information. Also, traditional convolution operation also computes slowly, so it should be replaced using a more efficient convolution operation. In addition, although the dense blocks in DenseNet have a good mobility and coupling for internal features, but the dense blocks and transition blocks in DenseNet are simply stacked together. As such, the output features of each dense block are not well exploited, e.g., hierarchical information of different layers are not fully utilized.

In this paper, we therefore propose effective strategies to resolve the aforementioned problems to improve the representation ability and efficiency of deep frameworks for text recognition. The contributions of this paper are summarized as follows:

1. A new framework called Fully-Convolutional Intensive Feature Flow Neural Network (IntensiveNet) is technically derived for recognizing texts in images. To enhance the feature flow and coupling between different levels of layers, we design a new further dense block, termed intensive block, where the input features and two dense modules are connected tightly. For our intensive block, we present the dense fusion block and further dense fusion operations to enhance the representation learning ability. By adding short connections to different layers, features of different layers can be fully utilized and IntensiveNet can potentially obtain higher performance in feature learning.

---

[1] School of Computer Science and Technology, Soochow University, China, emails: sdtzm2017@hotmail.com, cszzhang@gmail.com
[2] Key Laboratory of Knowledge Engineering with Big Data (Ministry of Education) & School of Computer and Information, Hefei University of Technology, Hefei, China, email: eric.mengwang@gmail.com
[3] Bio-Computing Research Center, Harbin Institute of Technology, Shenzhen 518055, China, email: darrenzz219@gmail.com
[4] Inception Institute of Artificial Intelligence, Abu Dhabi, UAE, email: qinjiebuaa@gmail.com

2. To potentially prevent important feature information being lost and make the parameters of the whole framework learnable at the same time, we use a convolution operation with stride 2 to replace the pooling operation as down-sampling strategy.

3. To improve the model efficiency, we use the depth-wise separable convolution to replace the standard convolution operation in the intensive block, because the depth-wise separable convolution can deliver comparable results to the standard convolution but using less computational cost.

The paper is outlined as follows. Section 2 introduces the related work. Section 3 presents our IntensiveNet. Section 4 shows the experimental results and analysis. In Section 5, we presents some discussions and remarks. The conclusion is given in Section 6.

## 2 RELATED WORK

### 2.1 Dense Convolutional Network (DenseNet)

CNNs obtain enhanced performance if there exists shorter connections between the layers close to input and those close to output [5]. To this end, a new CNN model called DenseNet [5] that connects each layer to other layers using a feed-forward fashion, is derived. Different from traditional models, the feature maps of all preceding layers are used as inputs and its own feature maps [21] are used as inputs into all subsequent layers in DenseNet [5]. Two important modules in DenseNet are dense block and transition layer.

**Dense block.** Inspired by ResNet that creates short paths between layers, a simple connectivity pattern called Dense block [5] was derived. To ensure maximum information flow between layers, all layers in a dense block are connected directly with each other [5]. To keep the feed-forward nature, each layer obtains additional inputs from all preceding layers and passes on its own features to all subsequent layers [5]. Fig. 1 shows the layout of a 3-layer dense block including an input layer and two convolutional layers, where the convolutional layer includes the functions of Batch Normalization (BN) [22], Rectified Liner Units (ReLU) [23] and Convolution. DenseNet combines the related features by concatenating them.

**Transition layer.** The concatenation operation used in dense connectivity [5] is not viable as the size of feature-maps changes. An essential part of the convolutional networks is down-sampling layer that changes the size of features [5]. Specifically, DenseNet divides the network into multiple dense blocks to facilitate down-sampling in the architecture and refers to the layers between dense blocks that include convolution and pooling as the transition layers.

### 2.2 Depth-wise Separable Convolution

The depth separable convolution is first proposed in MobileNet-V1 [17]. Different from the traditional convolution that considers both channel and region changes at the same time, the deep separable convolution achieves the separation of channel and region separation. Specifically, it divides the convolution operation into two sub-steps, i.e., depth-wise and point-wise processes.

Depth-wise is a process that divides the input features with the form of $N \times H \times W \times C$ into $C$ groups, and then performs $3 \times 3$ convolution operation to each group, where $N$ is the number of features, $H$ is the height of features, $W$ is the width of features and $C$ is the number of channels of features. The depth-wise process mainly collects the spatial features from each channel, namely, depth-wise features. The point-wise refers to the process that does $1 \times 1$ convolution operation using $k$ filters to the output features from depth-wise process, which collects features of each point, i.e., point-wise features. Fig.2 shows the depth separable convolution in the case of padding "same", where K denotes the convolution kernel. Note that although the output features of the depth separable convolution has the same size as those of the traditional convolution, it can significantly reduce the amount of parameters and computational cost.

### 2.3 Convolutional Recurrent Neural Network

CRNN is an end to end framework recognizing the text sequences in scenes [29]. As shown in Fig.3, CRNN has three parts. The first part is a Convolution layer that extracts features from images, the second is a Recurrent Layers that predicts the label distribution of each frame, and the last one is a Transcription layer that can transform the prediction of each frame into the final label sequences.

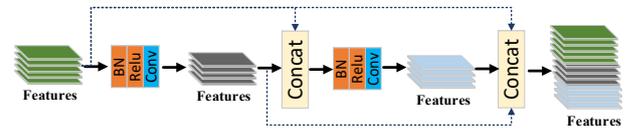

**Figure 1.** A 3-layer dense block with a growth rate being four.

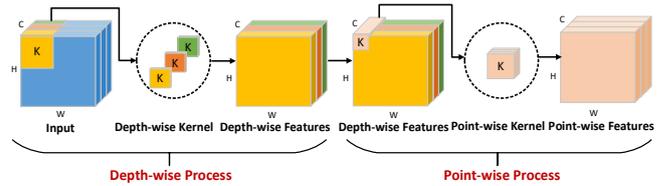

**Figure 2.** The process of depth separable convolution.

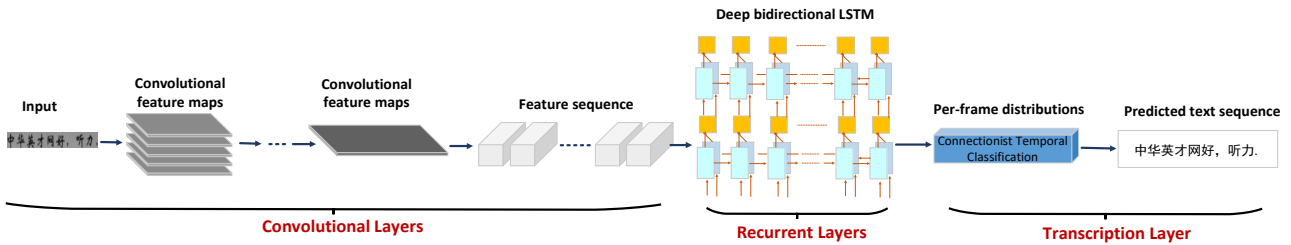

**Figure 3.** The learning architecture of CRNN for text recognition.

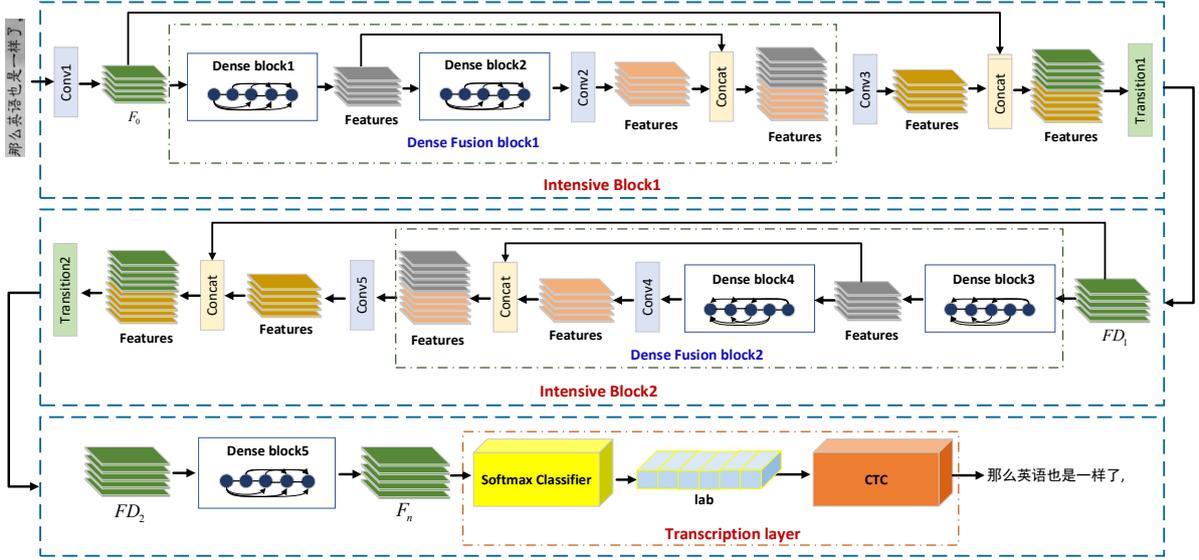

**Figure 4.** The learning architecture of our proposed framework for text recognition from images.

## 3 FULLY-CONVOLUTIONAL INTENSIVE FEATURE FLOW NEURAL NETWORK

In this section, we introduce the learning framework of our IntensiveNet. To be specific, we first describe the learning architecture of the whole framework of IntensiveNet, and then detail the proposed intensive block and transcription layer.

### 3.1 The Whole Framework

The whole architecture of IntensiveNet for the text and handwritten numbers recognition from images is shown in Fig. 4. We can find that our IntensiveNet is a fully-convolutional framework that mainly includes two intensive blocks and a transcription layer (soft-max and CTC). The first convolution operation, namely, Conv1, is used to extract shallow features. Moreover, Conv1 also plays the role of down-sampling when encountering with large-size features. This layer extracts features $F_0$ from input:

$$F_0 = H_{Conv1}(input) = H_{BN}(w \cdot H_{RELU}(input) + b), \quad (1)$$

where $H_{Conv1}(\bullet)$ is the separable convolution operations, $H_{BN}(\bullet)$ is Batch Normalization (BN), $w$ and $b$ are the parameters of weights and bias of convolution operations, $H_{RELU}(\bullet)$ is the Rectified Liner Units (ReLU) operation. All the separable convolution operations in this paper refer to an operation group that consists of ReLU, Depth separable convolution and BN, as shown in Fig. 5. Moreover, recent works find that the operation order of ReLU, convolution and BN may get better results. Thus, we adopt this strategy instead of the original operation order, i.e., BN, ReLU and convolution, in the framework of DenseNet. When features pass through Intensive Block1, we can obtain the high-mobility features $FD_1$:

$$FD_1 = H_{FDB1}(F_0), \quad (2)$$

where $H_{FDB1}(\bullet)$ is the function of Intensive Block1. Note that $FD_1$ contains the final features of Intensive Block1, which goes through the effects of the Dense Fusion block1 and several convolution and concatenating operations. Similarly, $FD_1$ are then fed into Intensive Block2. We can obtain features $FD_2$ after the Intensive Block2 as

$$FD_2 = H_{FDB2}(FD_1) = H_{FDB2}(H_{FDB1}(F_0)), \quad (3)$$

where $H_{FDB2}(\bullet)$ is the function of our Intensive Block2. Compared with original dense blocks, our features have better mobility and are more suitable for the feature fusion. More justifications will be provided in the later section. Finally, we use a dense block to extend the depth of our network to enhance the feature representation ability. We can obtain the final features $Fn$ as follows:

$$F_n = H_{dense5}(FD_2), \quad (4)$$

where $H_{dense5}(\bullet)$ is the function of dense block5. The features $F_n$ are then inputted into the soft-max classifier [3]. As such, we can obtain the predicted label $lab$ as follows:

$$lab = \text{soft-max}(F_n) = \sigma(z)_j = e^{z_j} / \sum_{k=1}^{K} e^{z_k}, \quad s.t. \ F_n = z, \quad (5)$$

where $\sigma(\bullet)$ is the soft-max function, $j$ is the $j$-th item in $F_n$, $K$ is the number of items in $F_n$. Note that we apply CTC to transform these predictions from classifier into the final label sequence in the problem of OCR. Next, the proposed intensive block will be described.

### 3.2 Our Intensive Block

We introduce the proposed intensive block. The proposed intensive block can enhance the feature flow and coupling between different layers, and it has two novel contributions, that is, (1) Dense Fusion block; (2) Further Dense Fusion operations.

**Dense Fusion block.** The Dense Fusion block consists of two dense blocks and a convolution operation. We take the Intensive Block1 as an example. According to Fig.5, $F_0$ is fed into the first dense block and obtain dense features $Fd_1$. Similarly, we can extract dense features $Fd_2$ from the second dense block as

$$\begin{aligned} Fd_2 &= H_{dense2}(Fd_1) \\ &= H_{concat}(Fd_1, Fd_{11}, Fd_{12}, ... Fd_{1c}) \\ &= [Fd_1, H_{conv}(Fd_1), H_{conv}(Fd_1 + H_{conv}(Fd_1)), ... \\ &\quad H_{conv}(Fd_1 + H_{conv}(Fd_1) + ... + H_{conv}(Fd_{1c-1}))] \end{aligned} \quad (6)$$

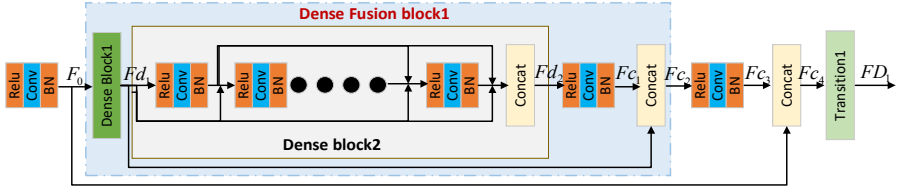

**Figure 5.** The structure of our proposed intensive block.

where $H_{dense2}(\bullet)$ is the function of dense block2, $H_{concat}(\bullet)$ is the function of concatenation, $Fd_{1i}$ $(i=1,2,...c)$ are inner features of each layer in dense block, $c$ is the number of layers in dense block, the operation $[\bullet,...,\bullet]$ means that features are concatenated in the direction of channels. The number of channels of features $Fd_2$ is defined as $N_{Fd_2} = N_{Fd_1} + c*g$, where $N_{Fd_1}$ is the number of channels of features $Fd_1$, $g$ is the growth rate. Then, we do a convolution operation to $Fd_2$ and obtain features $Fc_1$. This step extracts and learns features that are used to concatenate with features $Fd_1$ to enhance the mobility and fusion of dense blocks.

**Further Dense Fusion operations.** The remaining operations in the intensive block are called Further Dense Fusion operations. Similar to the operations in Dense Fusion block, we do a convolution operation to feature $Fc_2$ and concatenate $Fc_3$ with the original input $F_0$ to obtain combined features $Fc_4$. Finally, we use the transition block to down-sample $Fc_4$. The transition block is also a convolution operation, and it plays the role of the first convolution in Intensive Block2. By this way, we connect the adjacent intensive blocks tightly. Thus, we not only fully use the inner feature information of intensive block, but also enhance the mobility and fusion of global feature information for the whole framework. Note that one of our major contributions is to utilize several convolution and concatenating operations to construct the shortcuts for hierarchical features of different levels of layers. Hence, our network can fully excavate and utilize the features with different receptive fields [37]. Finally, we give the formula of intensive block as follows:

$$\begin{aligned}
FD_1 &= H_{trans}(Fc_4) \\
&= H_{trans}(H_{concat}(Fc_3, F_0)) \\
&= H_{trans}(H_{concat}(H_{conv}(Fc_2), F_0)) \\
&= H_{trans}(H_{concat}(H_{conv}(H_{concat}(Fc_1, Fd_1)), F_0)) \\
&= H_{trans}(H_{concat}(H_{conv}(H_{conv}(Fd_2), Fd_1)), F_0)) \\
&= H_{trans}(H_{concat}(H_{conv}(H_{conv}(H_{concat}(A)), Fd_1)), F_0))
\end{aligned} \quad (7)$$

where $A=(Fd_1, Fd_{11}, Fd_{12}, ...... Fd_{1c})$ denotes an auxiliary matrix.

### 3.3 Transcription Layer

This transcription layer is mainly used to transform the prediction of each frame into the final label sequence, which includes the soft-max and CTC. Soft-max is utilized to output the predictions of the last dense block. CTC plays the role in transforming these predictions into the final label sequence. In our proposed networks, CTC needs to input data of each column of a picture containing text as a sequence and outputs the corresponding characters.

## 4 EXPERIMENTAL RESULTS

In this section, we evaluate each method on two recognition tasks, i.e., recognizing texts in images and recognizing handwritten digits from images [38-40]. For the task of recognizing the texts in images, we compare the result of our framework with those of 6 popular deep models, where the CPUs and GPUs of all evaluated methods in experiments are Xeon E3 1230 and 1080 Ti respectively and the used convolution architecture is based on the framework of Caffe [31]. A popular large-scale synthesis Chinese String dataset [30] is used for evaluations. For the task of handwritten digits recognition from the images, we compare the recognition results with several popular methods on MNIST [15]. It is noteworthy to point out that CTC is applied in the task of text recognition but is not required in handwritten digits recognition. Because the task of handwritten recognition only needs to perform single character recognition, we utilize DenseNet to extract feature information and use soft-max [3] as the classifier directly to predict the labels of samples.

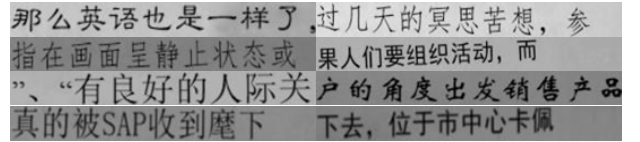

**Figure 6.** Illustration of image examples in Chinese string dataset.

**Table 1.** Comparison Results of Evaluated Deep Models on Synthetic Chinese String Dataset.

| Evaluated Frameworks | Accuracy |
|---|---|
| Inception-bn-res-blstm | 92.00 |
| Resnet-res-blstm | 91.00 |
| Densenet-res-blstm | 96.50 |
| Densenet-no-blstm | 97.00 |
| Densenet-sum-blstm-full-res-blstm | 98.05 |
| Densenet-no-blstm-vertical-feature | 98.16 |
| **Our IntensiveNet** | **98.67** |

**Table 2.** Comparison Results of Evaluated Deep Models on the MNIST Handwritten Digit Database.

| Evaluated Frameworks | Accuracy |
|---|---|
| Deep L2-SVM | 99.13 |
| Maxout Network | 99.06 |
| BinaryConnect | 98.71 |
| PCANet-1 | 99.38 |
| gcForest | 99.26 |
| Simple CNN with BaikalCMA loss | 99.47 |
| **Our IntensiveNet** | **99.73** |

## 4.1 Text Recognition in Images

In this section, we evaluate each deep network model for recognizing the texts in images using the Synthetic Chinese String dataset. The Chinese string data are generated randomly from the Chinese corpus, for instance news and classical Chinese, by changing fonts, sizes, gray levels, blurring, perspective and stretching, which is made by following the procedures in [30]. The dictionary consists of about 5990 characters, including Chinese, punctuation, English and numbers. Each sample is fixed to 10 characters, and characters are randomly intercepted from the corpus. The resolution of the pictures is unified to 280×32. A total of about 3 million 600 thousand images are generated, which are divided into training set and test set according to 9:1. Fig. 6 shows some image examples of the dataset. Note that the used dataset and compared methods are all publicly available at https://github.com/senlinuc/caffe_ocr.

**Implementation details.** In our experiments, we adopt the stochastic gradient descent (SGD) [28] for training the proposed deep network model. We take Tensorflow [32] and Keras as our experiment architectures. The training of the deep network is implemented on TITAN Xp. The batch size is set to 32 and the epoch size is 10. The initial learning rate is set to 0.001, which will be adjusted at each epoch with the algorithm of 0.005*0.4**epoch, where "**" denotes the power calculation. The weight decay is set to 0.0001. In our network, we also add a drop layer [27] after the last dense block and set the dropout rate to 0.2. We set growth rate and number of layers inside a dense block to (8, 8). We utilize the value of test loss as a monitor, and the process training will be stopped early when the loss value stops descending. The weights will be kept when the training of each epoch finishes.

**Recognition results.** We describe the comparison results in Table 1, where term "Accuracy" refers to the correct proportion of the whole string and statistics on the test set. For each evaluated compared method, the recognition results are based on the frameworks of CRNN/DenseNet plus CTC. The suffix "res-blstm" denotes the method with blstm [25] in the form of residuals, the suffix "no-blstm" means that there is no LSTM layer used in the frameworks. The framework of "Densenet-sum-blstm-full-res-blstm" has two changes over "Densenet-res-blstm": (1) the approach of combining two lstms into blstm changes from concat to sum; (2) both layers of blstm are connected using the residual way. "Densenet-no-blstm-vertical-feature" removes the pooling operations [26] of 1x4 relatively to "Densenet-no-blstm". The text recognition result of our model is obtained based on the intensive blocks. We can find that our IntensiveNet obtains the highest accuracy, compared with other related approaches, which implies that the proposed intensive block has played an important role in improving the recognition results.

## 4.2 Handwritten Digits Recognition from Images

We evaluate each deep framework for recognizing the handwritten digits based on images by using the popular MNIST database [15]. MNIST is a widely used dataset where the goal is to classify 28×28 pixel images as one of 10 digits. MNIST dataset has 60,000 training samples and 10,000 testing samples. The results on MNIST can reflect the ability of feature extraction and learning of a model.

**Implementation details.** For MNIST handwritten digit database, the batch size is set to 128 and the epoch size is 200. The initial learning rate is set to 0.001, which will be adjusted to 0.0001 at interval between 50 and 100, and to 0.00001 after the 100th epoch. In this simulation, the performance of our IntensiveNet is compared with those of six popular models, including Deep L2-SVM [41], Max-out Network [42], BinaryConnect [43], PCANet-1 [44], gcForest [45] and Simple CNN with BaikalCMA loss.

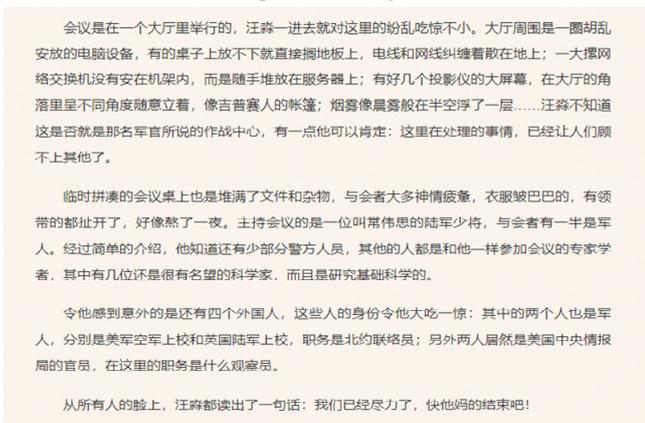
**Inputted text image**

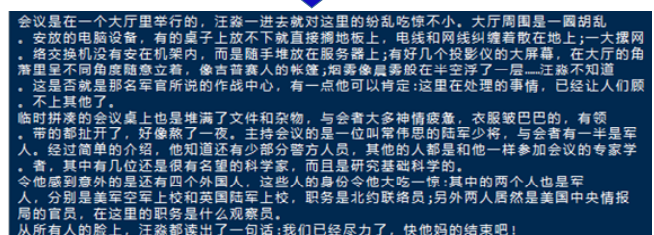
**Direct text output by our method**

**(a) Example one**

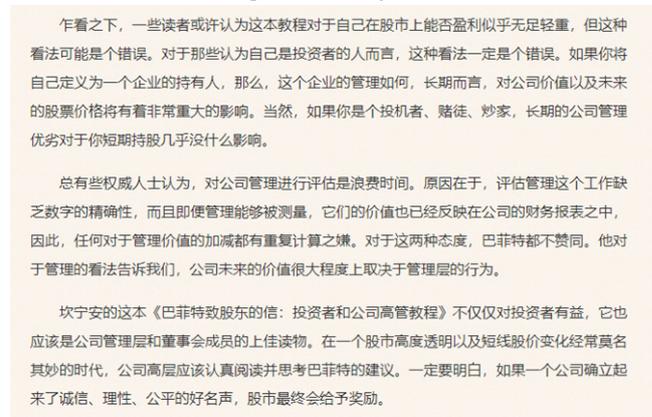
**Inputted text image**

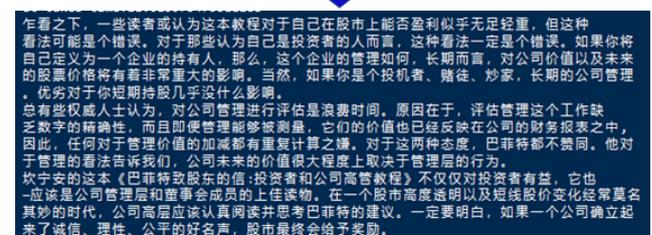
**Direct text output by our method**

**(b) Example two**

**Inputted text image**

我比现在年轻十几岁的时候，获得了一个游手好闲的职业，去乡间收集民间歌谣。那一年的
整个夏天，我如同一只乱飞的麻雀，游荡在知了和阳光充斥的农村。我喜欢喝农民那种带有苦
味的茶水，他们的茶桶就放在田埂的树下，我毫无顾忌地拿起积满茶垢的茶碗舀水喝，还把自
己的水壶灌满，与田里干活的男人说上几句废话，在姑娘因我而起的窃窃私笑里扬长而去。我
曾经和一位守着瓜田的老人聊了整整一个午后，这是我有生以来瓜吃得最多的一次，当我站起
来告辞时，突然发现自己像个孕妇一样步履艰难了。然后我与一位当上了祖母的女人坐在门槛
上，她编着草鞋为我唱了一支《十月怀胎》。我最喜欢的是傍晚来到时，坐在农民的屋前，看
着他们将提上的井水泼在地上，压住蒸腾的尘土，夕阳的光芒在树梢上照射下来，拿一把他们
递过来的扇子，尝尝他们的盐一样咸的咸菜，看着几个年轻女人，和男人们说着话。

我头戴宽边草帽，脚上穿着拖鞋，一条毛巾挂在身后的皮带上，让它像尾巴似的拍打着我
的屁股。我整日张大嘴巴打着哈欠，散漫地走在田间小道上，我的拖鞋吧嗒吧嗒，把那些小道
弄得尘土飞扬，仿佛是车轮滚滚而过时的情景。

我到处游荡，已经弄不清楚哪些村庄我曾经去过，哪些我没有去过。我走近一个村子时，
常会听到孩子的喊叫：

"那个打哈欠的人又来啦。"

↓ **Our deep network model**

我比现在年轻十几岁的时候，获得了一个游手好闲的职业，去乡间收集民间歌谣。那一年的
整个夏天，我如同一只乱飞的麻雀，游荡在知了和阳光充斥的农村。我喜欢喝农民那种带有苦
味的茶水，他们的茶桶就放在田埂的树下，我毫无顾忌地拿起积满茶垢的茶碗舀水喝，还把自
己的水壶灌满，与田里干活的男人说上几句废话，在姑娘因我而起的窃窃私笑里扬长而去。我
曾经和一位守着瓜田的老人聊了整整一个午后，这是我有生以来瓜吃得最多的一次，当我站起
来告辞时，突然发现自己像个孕妇一样步履艰难了。然后我与一位当上了祖母的女人坐在门槛
上，她编着草鞋为我唱了一支《十月怀胎》。我最喜欢的是傍晚来到时，坐在农民的屋前，看
着他们将提上的井水泼在地上，压住蒸腾的尘土，夕阳的光芒在树梢上照射下来，拿一把他们
递过来的扇子，尝尝他们的盐一样咸的咸菜，看着几个年轻女人，和男人们说着话。
，我头戴宽边草帽，脚上穿着拖鞋，一条毛巾挂在身后的皮带上，让它像尾巴似的拍打着我
的屁股。我整日张大嘴巴打着哈欠，散漫地走在田间小道上，我的拖鞋吧嗒吧嗒，把那些小道
弄得尘土飞扬，仿佛是车轮滚滚而过时的情景。
，我到处游荡，已经弄不清楚哪些村庄我曾经去过，哪些我没有去过。我走近一个村子时，
常会听到孩子的喊叫：
"那个打哈欠的人又来啦。"

**Direct text output by our method**

**(c) Example three**

**Inputted text image**

阳光洒洒。

叶钦戴着草帽，一手提着竹篮，沿着齐膝高的河水，缓缓走动着，双脚打破了河
面的平静，呼啦啦的水流声不时响起。

透过清澈见底的河水，叶钦的目光一直逡巡着河底的鹅卵石，不时地俯身，手臂
穿过温热的水流，在鹅卵石上摸索着。

每一次俯身他都会从鹅卵石上捡起几个螺蛳，随手扔进挎着的竹篮里。

螺蛳在南方的溪流小河里很普遍，形状很像田螺，不过它的贝壳表面不像田螺那
样光滑，而生长着许多旋转的肋纹，而且个头也多数偏小一些。

不过，蚊子腿再小也是肉呢。

叶钦看着竹篮里的螺蛳，盘算着回去大概能够炒上一碗了，轻轻咽了下口水。奶
奶厨艺很好，这些螺蛳拿回去用清水漂个一下午，晚上就能炒了。

今天早上就喝了两碗地瓜粥，当时还觉得有些胀肚子，跑出来转悠了一上午，出
点汗，动一动，就觉得自己这会有些饿了。

↓ **Our deep network model**

阳光洒洒。
叶钦戴着草帽，一手提着竹篮，沿着齐膝高的河水，缓缓走动着，双脚打破了河
面的平静，呼啦啦的水流声不时响起。
!透过清澈见底的河水，叶钦的目光一直逡巡着河底的鹅卵石，不时地俯身，手臂
穿过温热的水流，在鹅卵石上摸索着。
!每一次俯身他都会从鹅卵石上捡起几个螺蛳，随手扔进挎着的竹篮里。
螺啊在南方的溪流小河里很普遍，形状很像田螺，不过它的贝壳表面不像田螺那
!样光滑，而生长着许多旋转的肋纹，而且个头也多数偏小一些。
不过，蚊子腿再小也是肉呢。
叶钦看着竹篮里的螺蛳，盘算着回去大概能够炒上一碗了，轻轻咽了下口水。奶
奶厨艺很好，这些螺蛳拿回去用清水漂个一下午，晚上就能炒了。
今天早上就喝了两碗地瓜粥，当时还觉得有些胀肚子，跑出来转悠了一上午，出
点汗，动一动，就觉得自己这会有些饿了。

**Direct text output by our method**

**(d) Example four**

**Figure 7.** Illustration of some Chinese recognition results using our proposed IntensiveNet framework.

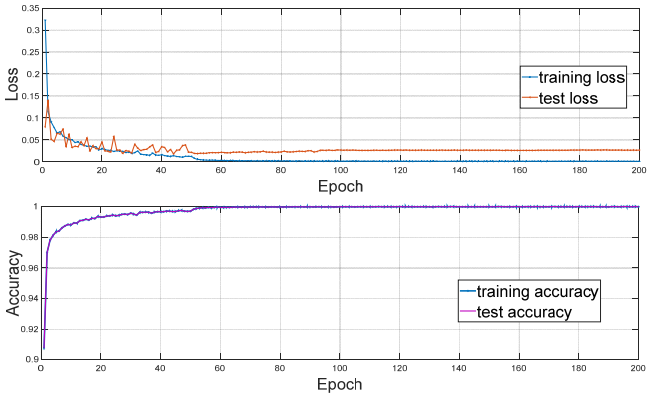

**Figure 8.** Training curves of handwritten digits recognition on MNIST.

**Recognition results.** The recognition result in terms of accuracy rate on MNIST is described in Table 2. We can see that our model can obtain the enhanced results, which can once again demonstrate that the proposed intensive feature flow neural network and according strategies of intensive blocks are effective.

### 4.3 Visualization of Recognized Texts in Images

In addition to the quantitative evaluation results, we also visualize some recognized texts in images by our model in Fig.7. To visualize the results, we utilize the CTPN method to extract the key text lines from test images. We can observe that our model can output high-quality text recognition results. For some identified sentences, there have some deviations in text positions, which is caused by the fact that CTPN has no layout analysis function making CTPN fail to produce accurate text alignment when detecting text lines.

### 4.4 Visualization of Training Process

In this experiment, we illustrate the training curve of our proposed IntensiveNet that is trained using the cross-entropy loss on MNIST handwritten digit database. The training results based on different numbers of epochs are shown in Fig. 8, where the top figure represents the curve of the cross-entropy loss and the bottom one shows the recognition accuracy. We can find that the ross-entropy loss is well fitted, and the curves of recognition accuracies of testing and training almost coincide, which implies that our IntensiveNet has a strong ability of feature representation and learning.

## 5 DISCUSSION AND SOME REMARKS

In this section, we discuss some important issues that are closely related to our proposed deep framework.

**Difference between DenseNet and IntensiveNet.** Inspired by DenseNet, we use several shortcuts in our proposed intensive block. In general, DenseNet is widely used in high-level computer vision tasks. Our method aims to solve the sub-tasks of image recognition. We replace the inefficient convolutional operation using the highly efficient depth separable convolution to reduce the amount of parameters and cost of computation. We also remove the pooling layers that may lose feature information and instead use the convolution with stride 2 as the strategy for down-sampling. So, IntensiveNet is a real fully-convolutional network. Also, we use several convolution and concatenating operations to connect the features from input and dense blocks, which could fully use the inner features from the intensive block. We take the transition layer in the previous intensive block as the first convolution operation of next intensive block. Thus, global features of the whole framework can be extracted and learned. By this way, we can make full use of the hierarchical features from different dense blocks and capture global features, both of which are neglected in DenseNet.

**About the first convolutional layer in IntensiveNet.** The first convolution operation in our model has different functions according to different inputs. For input images with small sizes, it aims at extracting shallow features, e.g., MNIST. But when facing large-size images, the first convolution usually set its stride as 2 to down-sample the original input. In this case, it will play the same role as the transition layers. But the difference is that the kernel size of the first convolution of the whole framework is 5*5 for bigger receptive field, not 3*3 in the other convolution operations. In theory, we can obtain better performances with deeper networks and input images with larger sizes that contain more feature information. But due to the limits of the computing power and resources, we have to use the down-sampling strategy to features with large sizes. In fact, we can magnify the original images by up-sampling such as transpose convolution. Under this circumstance, we can further extend the depth of network and the final result is expected to be improved.

**Two strategies to extend the depth.** In our framework, we utilize two dense blocks and several convolution operations to define the new intensive block. So as to extend the depth of our proposed network, two strategies can be used. One is adding more intensive blocks, and another one is adding more dense blocks in the Dense Fusion blocks. The former one can enhance the mobility and fusion of global feature information, while the latter aims at enhancing the mobility and fusion of the local feature information. In fact, we can also combine the two extending strategies. As such, we can obtain a network with better ability of feature leaning by fully combining features from different levels of layers.

**Discussion on the used datasets.** In this paper, we apply a Synthetic Chinese String dataset and MNIST for the tasks of texts and handwritten numbers recognition, respectively. These two tasks are all the processes of recognizing texts based on images, but the former tackles the task that many texts are in the same image while handwritten numbers recognition handles the task that each image contains only one digit or number. MNIST is the earliest and most popular handwritten digital dataset. Due to the simplicity of data distribution, many existing models can obtain encouraging results on it, including traditional shallow learning methods or CNNs with few layers. Notwithstanding the results on MNIST can also reflect the ability of feature extraction and learning of a model. The results of our IntensiveNet have shown its excellent ability. From the large Chinese string dataset whose distribution of samples is extensive and complex, IntensiveNet consistently achieves impressive performance than other deep models. In other words, our IntensiveNet has extraordinary potentials for the text recognition on both small-scale and large-scale datasets in the real applications.

## 6 CONCLUSION AND FUTURE WORK

We propose a novel deep model termed Fully-Convolutional Intensive Feature Flow Neural Network called IntensiveNet for Chinese and handwritten digits recognition. Our IntensiveNet improves the performance by preventing the loss of feature information due to pooling operation, enhancing the mobility and fusion of features from different layers, and improving the computational efficiency of the convolution operations further. Specifically, we propose the concept of intensive block that can enhance the mobility and fusion of hierarchical features from different layers. Besides, IntensiveNet applies the depth-wise separable convolution to make the process efficient, and uses a convolution operation with stride 2 to replace the pooling step in the customary transition layers, which makes the parameters of the whole network learnable and can also prevent the loss of feature information at the same time.

We examined IntensiveNet for the Chinese and handwritten digits recognition, and enhanced results are obtained by our method compared with related deep models. In future, more effective ways to enhance the feature flow and retain more important features are highly-desired to be explored. We will also explore how to choose the optimal number of layers in the deep frameworks, including our model. In addition, due to the promising performance of the fully-convolutional models, we will extend the proposed architecture to other related applications, e.g., object segmentation [33-35].

## ACKNOWLEDGEMENTS

This paper is partially supported by the National Natural Science Foundation of China (61672365, 61732008, 61725203, 61622305, 61871444 and 61806035), and the Fundamental Research Funds for the Central Universities of China (JZ2019HGPA0102). Zhao Zhang is the corresponding author of this paper.